Kris A. G. Wyckhuys
School of the Environment
University of Queensland
Saint Lucia, Australia
Contact: k.wyckhuys@uq.edu.au

# General-purpose AI models can generate actionable knowledge on agroecological crop protection

Kris A.G. Wyckhuys[1,2,3]

1. Chrysalis Consulting, Danang, Vietnam
2. Institute for Plant Protection, China Academy of Agricultural Sciences (CAAS), Beijing, China
3. School of the Environment, University of Queensland, Saint Lucia, Australia

## Abstract

Generative artificial intelligence (AI) offers potential for democratizing scientific knowledge and converting this to clear, actionable information, yet its application in agri-food science remains unexplored. Here, we verify the scientific knowledge on agroecological crop protection that is generated by either web-grounded or non-grounded large language models (LLMs), i.e., DeepSeek versus the free-tier version of ChatGPT. For nine globally limiting pests, weeds, and plant diseases, we assessed the factual accuracy, data consistency, and breadth of knowledge or data completeness of each LLM. Overall, DeepSeek consistently screened a 4.8-49.7-fold larger literature corpus and reported 1.6-2.4-fold more biological control agents or management solutions than ChatGPT. As a result, DeepSeek reported 21.6% higher efficacy estimates, exhibited greater laboratory-to-field data consistency, and showed more realistic effects of pest identity and management tactics. However, both models hallucinated, i.e., fabricated fictitious agents or references, reported on implausible ecological interactions or outcomes, confused scientific nomenclatures, and omitted data on key agents or solutions. Despite these shortcomings, both LLMs correctly reported low-resolution efficacy trends. Overall, when paired with rigorous human oversight, LLMs could support farm-level decision-making and unleash scientific creativity.

## Introduction

Over the past five years, generative artificial intelligence (AI) has seen a widespread proliferation. Large language models (LLMs) now exert disruptive impacts on diverse domains such as healthcare, data management, education and manufacturing, with potentially profound societal implications (Bommasani et al., 2021; Baidoo-Anu & Ansah, 2023). As self-supervised machine learning systems that autonomously identify, extract and learn from data, two LLMs in particular have earned global acclaim (Jin et al., 2025). ChatGPT (Generative Pre-trained Transformer) is a commercial, closed-source LLM that is valued for its user-friendliness, multi-modal capability and conversational abilities (OpenAI, 2024). Up till mid-2015, ChatGPT acted as a non-grounded model unable to produce outputs beyond what featured in its initial training datasets. Meanwhile, DeepSeek is a recently launched, open-source model with structured, human-like reasoning capabilities and a web-grounded modality i.e., the ability to access the internet in real time (Conroy & Mallapaty, 2025; Normile, 2025). Its step-by-step reinforcement learning makes DeepSeek adept at solving technical problems and potentially contributing to science and decision-support (Gibney, 2025).

ChatGPT and DeepSeek carry prominent strengths and weaknesses, and have reached notable achievements especially in human medicine. ChatGPT exhibits expert proficiency at medical licensing exams or in summarizing electronic health records (Kung et al., 2023; Van Veen et al., 2024), whereas DeepSeek shows superior reasoning in clinical decision making (Tordjman et al., 2025). Meanwhile, both of these AI engines or chatbots perform similarly when diagnosing oral pathologies (Kaygisiz & Teke, 2025) or when answering radiotherapy questions (Luo et al., 2025). Yet, while both LLMs can advance science and practice across myriad fields, a comparative evaluation of their performance is rarely conducted beyond the confines of the medical domain.

Amongst its many advantages, generative AI could offload mundane, resource-intensive or repetitive tasks (Jin et al., 2023; Van Veen et al., 2024) and thereby free scientists' time for more strategic work. LLMs, for instance, hold the potential to streamline systematic literature reviews (SLRs) or assist with the decision-making of various stakeholders. In particular, SLR development is costly, labor-intensive and impeded by an explosive growth in publication volume (Bolaños et al., 2024; Scherbakov et al., 2025). Reliant upon manual keyword-based filtering, even domain experts now miss up to 13% of relevant papers (Gartlehner et al., 2020). Generative AI could help scientists and librarians 'cut through the clutter' (Joos et al., 2024) and reduce their workload by up to 88-98% (Scherbakov et al., 2025). Looking beyond SLRs, AI tools can also democratize scientific knowledge and assist the decision-making of myriad other stakeholders (Capraro et al., 2024). Yet, to what extent this knowledge has been incorporated in the trained models of non-grounded models or can be accessed by web-grounded ones is unknown. Further, it remains unclear whether the open-access versions of general-purpose models can produce decision-relevant information. This may also be relevant for fields such as agriculture - where the potential of generative AI has been understudied.

Generative AI is steadily but discreetly infiltrating agricultural science and practice, demonstrating potential to improve productivity, accelerate innovation and advance policies (De Clercq et al., 2024). LLM-powered chatbots or expert systems such as Norm, Kissan GPT or Bayer's GenAI already deliver on-demand agronomic expertise and personalized advice (Silva et al., 2023; Martson, 2023), offering unprecedented opportunities but also risks for sustainable agriculture. A GPT-powered Virtual Agronomist for example raises Kenyan farmers' yields 1.4- to 1.9-fold, costing merely US$ 1.50 per season (Shepherd et al., 2025). However, the platform promotes insecticide-coated seeds or overhead sprays with neonicotinoids, i.e., measures that divert notably from global best practice. Further, LLMs such as ChatGPT are used to identify the appropriate time for pest management interventions (Yang et al., 2024), forecast risk of plant diseases (Calone et al., 2025), fast-track pesticidal synthesis or steer breeding programs (Farmer et al., 2025). While AI-augmented R&D may already be the new reality, the extent to which the 'as is' version of general-purpose LLMs can inform decision-making by farmers and/or scientists remains unexplored.

An AI-enabled step-change is especially opportune in the domain of sustainable pest management. Animal pests, weeds and plant pathogens, responsible for 40% of on-farm losses (Oerke, 2006), are increasingly managed using chemical pesticides (Shattuck et al., 2023). Pesticides are now ubiquitous in open fields and protected horticulture, driving biodiversity loss, degrading ecosystem functioning and imperiling human health (Tang et al., 2025). As pesticide-centered management often proves unwarranted or economically unsound (Janssen & van Rijn, 2021; Perrot et al., 2025), scientists have been pursuing alternative forms of crop protection that rely on invertebrate biological control agents, microbials or sound agronomy. Alone or in combination, these non-chemical solutions can sustain food or feed security without jeopardizing human or environmental health (Deguine et al., 2023; Schneider et al., 2023; Schaffner et al., 2024). Yet, farmers are often unaware of or misinformed about the potential of non-chemical or agroecological crop protection, as concrete evidence is often locked behind paywalls, conveyed in jargon-laden scientific texts, or distorted by agrochemical retailers and crop consultants (Struelens et al., 2022). Generative AI may unlock this knowledge and guide (farmer) decision making on nature-friendly crop protection (Gohr et al., 2025), but its potential so far remains unexplored.

This study had two aims. First, we verified the overall scientific knowledge on non-chemical crop protection that can readily be accessed by the free-tier versions of web-grounded and non-grounded AI models. Second, we systematically compared factual accuracy, data consistency and breadth of knowledge generated by either model. For nine globally-limiting biotic stressors, we further compared the above parameters for queries that were restricted to particular geographies. Though exploratory, our evaluation is critical to ensuring responsible, LLM-driven decision support on sustainable crop protection.

## Materials & Methods

Overall, our work aimed to assess the extent to which common general-purpose LLMs have been trained on up-to-date scientific literature (non-grounded models) or are able to readily access this information from the web (web-grounded models). For non-grounded models, we thus assumed that all or most scientific articles and sources were already in the corpus of training data as foundational models. Specifically, we assessed the accuracy, consistency and completeness of two LLMs or AI search engines, i.e., ChatCGPT-4o (CG; OpenAI, San Francisco, USA) with a cut-off date of April 2025 and DeepSeek-R1 (DS; DeepSeek AI, Hangzhou, China), when subject to queries or prompts on non-chemical pest management. First, we submitted standardized AI queries to the freeware, i.e., non-subscription versions of both engines between June 20, 2025 and July 5, 2025. We did not use academic search engines such as ScholarAI - which offer limited search capabilities for non-subscribers. Prompts were specifically designed to assess management alternatives for nine globally-limiting agricultural insect pests, plant diseases and weeds (collectively: 'pests'). Insect pests included the silverleaf whitefly *Bemisia tabaci*, cotton bollworm or corn earworm *Helicoverpa armigera*, and diamondback moth *Plutella xylostella*. Diseases included potato late blight caused by *Phytophthora infestans*, wheat rust caused by *Puccinia* spp. and Fusarium head blight caused by *Fusarium* spp. Weeds included barnyard grasses *Echinochloa* spp., horseweed *Erigeron canadensis* and amaranths *Amaranthus* spp. For the three insect pests, we aimed to document the efficacy of five management tactics, i.e., insect-killing microorganisms, invertebrate predators, parasitoids, botanical extracts and (other) agroecological measures including cultural control. For plant diseases and weeds, we only considered antagonistic or plant pathogenic microorganisms, botanical extracts and agroecological measures.

 Overall, we aimed to document efficacy of either management tactic against each pest at the global level and also for specific countries, i.e., China, Indonesia and Thailand. We intended to gauge the efficacy for each pest-tactic combination under laboratory, greenhouse or screenhouse, and field conditions. For insect pests, efficacy either captured the percentage reduction in pest density post-treatment (microorganisms, botanicals or agroecological measures), percentage reduction of pest populations, e.g., based on the number of nymphs and/or adults consumed (predators) or parasitism level (parasitoids). Meanwhile, efficacy captured the percentage reduction in disease incidence post-treatment for plant diseases or the percentage reduction in weed density post-treatment for weeds. Efficacy data were reported as mean values ± SD and the corresponding range. For efficacy data that pertained to a given country, the search engine was instructed to solely consider publications when they contained original data on tactic efficacy under laboratory, greenhouse, or field conditions, based on trials that had either been partially or entirely conducted in the study country.

Each standardized AI prompt or query conveyed detailed instructions on literature screening, data retrieval, analysis and presentation with a well-defined set of inclusion and exclusion criteria (Supplementary Table 1). In brief, prompts were defined so as to screen ISI peer-reviewed publications in any language that addressed the efficacy of specific non-chemical management tactics against a given insect pest, weed or plant disease in a particular geography. Such standardized, elaborate and carefully structured instructions are especially important given the engines' varying sensitivity to precise prompting (Jin et al., 2025). For each given pest, tactic or geography, a detailed prompt was obtained by initially submitting a concise, one-phrase query to DS and subsequently asking it to outline its literature review procedure. The ensuing self-reported procedure was then carefully revised, adapted to other pests, tactics and geographies and submitted as a formal query to either of the two search engines. For a select set of queries and taxa within restricted geographies, each engine was also tasked with listing the underlying scientific publications. This was aimed at verifying the validity of publication sources.

Next, AI-reported laboratory and field efficacy data (i.e., means and maxima) were tabulated per tactic, target pest and geography. To assess reporting accuracy, we systematically compared AI-reported efficacy data for each pest-tactic combination between the two AI engines. We also determined reporting internal consistency for either AI engine by contrasting its respective laboratory and field efficacy data for specific management options or agents, and exploring the underlying determinants (i.e., pest identity or management tactic). Meanwhile, we assessed external consistency by regressing DS-reported field efficacy data of specific solutions against the respective values reported by CG.

Lastly, we gauged the completeness of AI reporting by comparing the total number of listed agents or solutions and the total number of consulted scientific publications by either DS or CG for each pest-tactic combination. A sub-set of outputs (i.e., *B. tabaci* management in China) were complemented by asking both AI engines to disclose their underlying literature review process through Preferred Reporting Items for Systematic Reviews and Meta-Analyses (PRISMA). Specifically, we prompted each engine to list the literature resources that were consulted and to draw up a detailed PRISMA flow diagram. For each step of this diagram, e.g., number of initial records identified, records excluded at the abstract level, we then computed the disparity in data that were provided by either engine. Data disparity was expressed as the relative or percentual difference (%) in the number of records screened, excluded or included by either search engine.

*Data analysis*

For each pest, Analysis of Variance (ANOVA) was used to compare efficacy data between management tactics and AI engines. Further, linear regression analysis was used to relate field and laboratory efficacy values, as reported by either DS or CG for specific management solutions or agents. Eventual effects of pest identity or tactic were determined by entering these variables into the model in a stepwise fashion. Linear regression analysis was performed to relate DS-reported field efficacy for specific solutions or agents with the corresponding CG values. The number of listed solutions or agents and the number of consulted publications per tactic by each engine was compared using a paired samples t-test. The t-test was also used to compare field and laboratory efficacy data as reported by either engine. All data were checked for normality and homoscedasticity prior to analysis, and non-parametric tests were used for data that did meet normality assumptions. For all statistical analyses, IBM SPSS 30.0 was used.

## Results

*Insect pest management efficacy*

For laboratory data, as reported in the global literature, AI-reported maximum efficacy of *B. tabaci* management tactics differed between tactics (ANOVA, $F_{4,81}$= 9.261, p< 0.001), search engines ($F_{1,81}$= 89.651, p< 0.001) and the respective interaction term ($F_{4,81}$= 14.425, p< 0.001; Fig. 1A). Efficacy was highest for microbials (mean ± SD; 88.6 ± 4.8%) and lowest for agroecological methods (80.6 ± 14.9%). Across tactics, DS consistently reported higher efficacy levels (90.0 ± 4.5%) than CG (75.5 ± 16.8%; Fig. 1A).

For field data, on the other hand, maximum efficacy data for this pest was comparable between the two search engines ($F_{1,75}$= 0.961, p= 0.330) while differing between tactics ($F_{4,75}$= 7.926, p< 0.001) and the interaction term ($F_{4,75}$= 9.310, p< 0.001; Fig. 1B). Efficacy was highest for botanicals (77.6 ± 7.9), and lowest for parasitoids (67.2 ± 11.5). Across tactics, DS and CG reported data for a respective 13.8 ± 2.8 and 5.8 ± 1.3 agents, extracts or methods, as drawn from a respective 774.4 ± 536.1 and 17.4 ± 8.6 publications. The literature base for CG reports thus proved 97.8% smaller than that of DS.

In laboratory studies of *H. armigera* (Fig 1A), AI-reported maximum laboratory efficacy differed between tactics ($F_{4,105}$= 9.122, p< 0.001), engines ($F_{1,105}$= 51.672, p< 0.001) and the respective interaction term ($F_{4,105}$= 7.381, p< 0.001). Efficacy was highest for agroecological methods (91.3 ± 5.7%), and lowest for parasitoids (84.8 ± 10.9%). Across tactics, DS consistently reported higher efficacy levels (92.6 ± 4.7%) than CG (84.0 ± 8.7%).

In field trials, maximum field efficacy levels for this pest (Fig. 1B) differed between tactics ($F_{4,108}$= 15.316, p< 0.001), engines ($F_{1,108}$= 156.708, p< 0.001) and the interaction term ($F_{4,108}$= 35.615, p< 0.001). Microbials and parasitoids exhibited either the highest or lowest field efficacy (79.1 ± 8.9% and 69.4 ± 14.5%, respectively). As above, DS reported notably higher efficacy levels (81.6 ± 7.5) than CG (63.2 ± 12.9). Across tactics, DS and CG reported data for a respective 14.8 ± 3.8 and 9.0 ± 2.2 agents, extracts or methods, as drawn from a respective 2,668.6 ± 1,412.5 and 550.8 ± 632.0 publications. As such, the literature base for CG reports was 79.4% smaller than for DS, which may explain the (seemingly) greater accuracy for DS.

In laboratory studies of *P. xylostella*, maximum laboratory efficacy differed between tactics ($F_{4,67}$= 3.101, p= 0.021), engines ($F_{1,67}$= 86.787, p< 0.001) and the respective interaction term ($F_{4,67}$= 2.950, p= 0.026) (Fig. 3A). When tested in the laboratory, predators exhibited the highest efficacy (89.1 ± 8.1%) whereas agroecological measures proved least efficacious (70.2 ± 22.4%). Across tactics, DS reported markedly higher efficacy levels (91.5 ± 5.5%) than CG (69.4 ± 16.2%) (Fig 1A).

In field studies of *P. xylostella*, maximum efficacy also differed between tactics ($F_{4,64}$= 9.868, p< 0.001), engines ($F_{1,64}$= 24.295, p< 0.001) and a tactic x engine interaction term ($F_{4,64}$= 9.405, p< 0.001) (Fig. 3B). Under field conditions, efficacy was highest for predators (84.6 ± 6.3) and lowest for botanical insecticides (73.7 ± 17.5). As above, DS reported higher efficacy levels (80.7 ± 8.5) than CG (57.3 ± 21.5). Across tactics, DS and CG reported data for a respective 11.0 ± 1.0 and 6.2 ± 2.2 agents, extracts or methods, drawn from a respective 1,938.8 ± 756.7 and 39.0 ± 18.6 publications. As above, the literature base for CG reports was substantially (98.0%) smaller than for DS.

Overall, per pest and management tactic, DeepSeek consistently reported more solutions or biological control agents (Paired samples t-test, t= 5.385, df= 14, p< 0.001) and consulted more publications (t= 6.001, df= 14, p< 0.001) than ChatGPT. For all three pests, similar patterns were detected for each of the individual countries. Yet, country-level analyses were regularly (though not always) built upon a smaller literature base. For *B. tabaci* specifically, country-level reports for China, Thailand and Indonesia were built on a respective 11.6%, 6.8% and 6.6% (DeepSeek) and surprisingly 122%, 16.1% and 3.4% (ChatGPT) of the global literature bodies for either engine i.e., as reported above.

*Data consistency*

Across pests and tactics, AI-reported field efficacy estimates were consistent with their corresponding laboratory values for DeepSeek (averages: $F_{1,185}$= 960.007, p< 0.001, $R^2$= 0.838; maxima: $F_{1,185}$= 790.948, p< 0.001, R2= 0.810). Similar overall trends were observed for ChatGPT, though with markedly lower explanatory power (averages: $F_{1,71}$= 19.928, p< 0.001, $R^2$= 0.219; maxima: $F_{1,71}$= 11.037, p< 0.001, $R^2$= 0.135). DS-reported average or maximum field efficacy data were a respective 15.8-25.9% and 7.8-15.6% lower than their corresponding laboratory values (Paired samples t-test; t= 72.037, df= 186, p< 0.001; t= 37.836, df= 186, p< 0.001, respectively). Meanwhile, CG-reported average or maximum field efficacy data were a respective 3.9-22.8% and 2.3-23.4% lower than their corresponding laboratory values (t= 8.316, df= 72, p< 0.001; t= 7.190, df= 72, p< 0.001), exhibiting higher variability in laboratory-to-field differences.

For DS-reported average values, field efficacy could be explained by multivariate regression models ($F_{3,183}$= 403.798, p<0.001, $R^2$= 0.869) with statistically significant effects for laboratory efficacy (t= 30.075, p< 0.001), pest identity (t= 4.308, p< 0.001) and management tactic (t= 4.963, p< 0.001) (Fig. 2). Equally, for DS-reported maximum values, field efficacy could also be explained through linear regression ($F_{3,183}$= 610.553,

p< 0.001, $R^2$= 0.909) with significant effects of laboratory efficacy (t= 33.441, p< 0.001), pest identity (t= 12.555, p< 0.001) and management tactic (t= 6.540, p< 0.001). On the other hand, multivariate regression explained CG-reported average field efficacy ($F_{2,70}$= 14.643, p< 0.001, $R^2$= 0.295) with weaker effects for laboratory efficacy (t= 3.279, p= 0.002) and pest identity (t= -2.743, p= 0.008). Lastly, for CG-reported maximum efficacy, pest identity (t= -1.993, p= 0.05) or management tactic (t= -1.431, p= 0.157) did not improve overall significance of linear regression models.

Across tactics, the average field efficacy of individual agents or solutions reported by DS was inconsistent with the one reported CG ($F_{1,39}$= 1.028, p= 0.317, $R^2$= 0.026). Similarly, DS-reported maximum efficacy values were not in line with those reported by CG ($F_{1,38}$= 1.038, p= 0.315, $R^2$= 0.027) (Fig. 3).

*Disease and weed management efficacy*

Given the high reporting and data consistency in DeepSeek, we solely relied upon this search engine to assess disease management efficacy. At both global and national levels, botanicals and agroecology measures slightly outperformed microbial fungicides or bactericides under field conditions (Table 2). Specifically, AI consistently identified botanical mixtures and a combination of agroecological preventative measures as the most efficacious means (i.e., efficacy levels of 90-93%) to manage potato blight, wheat rust or Fusarium head blight. For each of the above tactics and diseases, efficacy data for China were comparable to those reported at the global level. In contrast, for management of the globally important *Echinochloa* spp. and *Erigeron canadensis*, mixtures of weed-killing microorganisms reported proved most efficacious (Table 3). Field applications of these agents attained efficacy levels up to 95% against either weed species. Meanwhile, for management of *Amaranthus* spp., botanical mixtures and allelopathic intercrops offered a small edge over microorganisms – attaining efficacy levels up to 93-95% in the field (Table 2). As above, across weeds and management tactics, the efficacy data for China differed only slightly from those reported at the global level.

As above, DeepSeek systematically screened a vast corpus of literature and provided efficacy data for multiple agents or solutions. Across the three disease targets, DeepSeek reportedly extracted efficacy data for 6.9 ± 0.6 agents from no less than 435.6

± 167.1 publications. Similarly, for either of the three weed targets, efficacy data were reported for 6.0 ± 0.9 agents as drawn from 314.1 ± 106.2 publications.

*AI literature review process*

When restricting searches to the Chinese *B. tabaci* literature, the initial and final number of publication records considered by either engine (marginally) differed (Mann-Whitney U= 3.00, P= 0.056; U= 0.00, p= 0.008, respectively). Specifically, while DS considered a respective 837.4 ± 303.3 and 135.6 ± 31.2 initial or final records, CG considered just 381.0 ± 359.5 and 22.6 ± 23.9 records. For the final number of studies that were considered, data disparity between both engines was as high as 172-175% for microbials or agroecological solutions (Table 1). As such, across tactics, ChatGPT's results were based upon data from 83% fewer literature sources than DeepSeek.

Throughout the literature retrieval, screening and analysis process, data disparity between DS and CG differed among the five pest management tactics (ANOVA, $F_{4,30}$= 6.947, p< 0.001; Table 1), with substantially lower values for invertebrate biological control agents. Data disparity was as low as 0.41 ± 0.98 and 0.44 ± 0.77 for predators or parasitoids, respectively, as compared to 1.53 ± 0.17 for microbials. Specifically, a comparable number of records were initially identified by either engine for invertebrate biological control agents i.e., with disparity levels of merely 12-17% (Table 1). Meanwhile, data disparity between either search engine did not differ among the seven successive steps of the PRISMA flow diagram ($F_{6,28}$= 0.964, p= 0.467). In addition, DS regularly screened a broader set of literature resources, covering Chinese language databases such as the Wanfang Data platform or CQVIP, and national repositories (Table 1).

## Discussion

By democratizing scientific knowledge, AI chatbots could support the decision-making of non-scientist actors and thereby assume a game-changing role in transforming today's agri-food systems (Candel, 2022; Schneider et al., 2023). In particular, AI tools could analyze, interpret, and synthesize vast bodies of digitized text and convert this into layman's terms almost instantaneously. Here, centering on sustainable pest management, we show how web-grounded models (DeepSeek) covered a 4.8-49.7-fold larger corpus of literature and reported 1.6-2.4-fold more biological control agents or solutions than non-grounded ones i.e., ChatGPT. The chatbot itself thus dictated efficacy metrics; GPT-reported (lab or field) data were 16.7-17.2% lower than those of DeepSeek. Data from GPT also had higher variability. Peak efficacy averaged 80.6-91.3% (laboratory) or 67.2-84.6% (field) across tactics, with the performance of key agents or solutions aligning with empirical data - thus affirming factual accuracy. Similar patterns were reported for weed and disease management, with multiple microbial, botanical or agroecological solutions providing more than 85-90% target control under field conditions. DeepSeek showed notably higher laboratory-to-field consistency for agent efficacy than ChatGPT, with more realistic pest- and tactic-mediated effects. Lastly, at the level of individual agents or solutions, the two models reported inconsistent peak and average field efficacy values. Overall, though model performance lacks validation against formal knowledge syntheses and human-run reviews (Khraisha et al., 2024), DeepSeek showed notable advantages in breadth of knowledge, data consistency and (seemingly) factual accuracy. Meanwhile, both models correctly identified key agents or solutions, and accurately picked up general performance trends. This bodes well for future human-machine collaboration in widening the reach of scientific knowledge in the agri-food domain.

Whereas ChatGPT and DeepSeek performed equivalently in screening, filtering, and synthesizing vast volumes of literature (Kaygisiz & Teke, 2025; Luo et al., 2025), the free-tier version of GPT4o fell short of expectations in our exploratory analysis. Its outputs were imbalanced, lacked precision and were almost generated by random-guessing or chance agreement (Khraisha et al., 2024). Further, even with elaborate, structured prompts, ChatGPT only identified or retained a respective 45.5% and 16.7% of the initial or final literature records that were processed by DeepSeek. Although both engines screened equivalent sets of databases (Table 1), the poor performance of ChatGPT4.o is unexpected (Mostafapour et al., 2024). This possibly can be attributed to three non-exclusive factors. First, GPT's free-tier version possibly shifts down to its older 3.5 model when faced with complex queries or extensive data retrieval tasks. ChatGPT's outputs, which hinted at restricted ISI literature database access, sporadically even conveyed hypothetical (instead of real) data on agent efficacy. Second, as LLM performance is mediated by language and domain expertise (Jin et al., 2025; Luo et al., 2025; Yuan et al., 2025), DeepSeek's proficiency in Chinese or English text analysis, structured problem solving and technical reasoning possibly gives it an edge in agricultural data synthesis. This may even become more pronounced as Chinese scientists already generate 60-65% of the world's agriculture or ecology papers and their outputs continue to rise (Anonymous, 2024; Tollefson, 2025). Third, ChatGPT is known to suffer from hallucination - the generation of fictitious content and false references (Emsley, 2023; Kacena et al., 2024; Susjnak et al., 2025; Bolaños et al., 2025) - but these issues may also affect DeepSeek, which could possibly explain its vastly superior literature coverage. Collectively, all three factors limit the usability of either or both LLMs for knowledge synthesis. More so, if the disparity in AI outputs is due to GPT's non-subscription mode, it could further widen the 'digital divide' with scientists and other stakeholders in low-income countries (De Clercq et al., 2024).

Beyond the mere size of the literature corpus, there are discrepancies in the actual content of AI-generated syntheses. First, both bots periodically listed implausible ecological interactions and impacts. Either bot reported up to 65-87% or 65-85% *B. tabaci* control efficacy (lab, field) by *Nosema bombycis* and *Spodoptera exigua* nucleopolyhedrosis virus (NPV), respectively, while neither of the two microorganisms has ever been reported affecting this pest (Suraporn & Terenius, 2021; Zhou et al., 2023). DeepSeek also 'hallucinated' non-existent agents – and associated references - such as *B. tabaci* NPV whereas no DNA viruses are known from this pest or any other members of the Aleyrodoidea (Gousi et al., 2025). Second, both bots reported on agroecological solutions that were impossible to implement in a laboratory. For instance, reduced tillage (ChatGPT) and hedgerow deployment (DeepSeek) reportedly suppressed *H. armigera* populations by a respective 75-80% and 85-87% in the laboratory.

Third, DeepSeek poorly distinguished between old and new nomenclature of biological control agents e.g., reporting *Paecilomyces fumosoroseus* (old; 15 papers) and *Isaria fumosorosea* (new; 87 papers) as two distinct species. Fourth, while LLM outputs appeared little skewed by dataset imbalance, i.e., overly dominated by the majority class (Khraisha et al., 2024), both bots picked up on rare records. Amongst a corpus of 18 *B. tabaci* microbial biopesticide publications, ChatGPT reported the performance of *Cordyceps javanica* (2 papers), *Aschersonia aleyrodis* (2) or *Purpureocillium lilacinum* (1) – neither of which featured in DeepSeek's corpus of 688 publications. Conversely, the latter engine reported on, e.g., *Akanthomyces lecanii* – which was not covered by GPT. Lastly, its narrower knowledge base sporadically led ChatGPT to dismiss the role of entire guilds: at the global level, *P. xylostella* field-level control efficacy was reported for only one microorganism i.e., *Bacillus thuringiensis* (Bt) whereas invertebrate predators were omitted altogether. Even though data are sparse, especially for predation (Furlong et al., 2013), there are credible records outlining the performance of invertebrates and microbiota besides Bt which cannot be overlooked (e.g., Quan et al., 2011; Hosseini et al., 2012). This kind of erroneous data omission has also been identified in the medical domain (Chen et al., 2025).

The above issues also plague LLM output on weed or disease management (Table 2, 3). No information could be found on the degree of *C. canadensis* control by *Alternaria alternata*, but other fungi such as *Albifimbria verrucaria* achieve 90-100% weed control (Hoagland et al., 2023) whereas *Xanthomonas campestris* causes 80% *C. canadensis* mortality in the greenhouse (Boyette & Hoagland, 2015). Similarly, while *Bacillus velezensis* provides up to 86% *P. infestans* control (Kim et al., 2021) and *Trichoderma harzianum* offers 90-98% control in the field, outperforming chemical fungicides (Mollah & Hassan, 2023), no published records were found on the superior efficacy of *Pseudomonas fluorescens* or *Paenibacillus polymyxa* reported by DeepSeek. Also, while turmeric extract does act against several *Fusarium* spp. (Avanço et al., 2017; Alsahli et al., 2018) and wheat rust (Kim et al., 2003), we could not corroborate the AI-claimed high field efficacy against the other two pathogens in our study. Evidently, these faulty outputs constrain the utility of either LLM, and will hopefully be resolved as AI-powered research assistants emerge (Bolaños et al., 2024).

Regardless of the above, both LLMs accurately portrayed non-chemical measures as valid pesticide alternatives. AI-reported peak performance of microbial biopesticides against *B. tabaci* (71%) remained within the 60-75% range achievable in the early 2000s (Faria & Wright, 2001). The AI-reported laboratory efficacy of *P. fumosoroseus* (72.8 ± 9.2%; range 50-88%) was even slightly inferior to the 68-94% efficacy reported by Vidal et al. (1997a), whereas the respective peak efficacy of 83% or 78% for *C. javanica* and *Beauveria bassiana* was below newly published values of 89-97% (Cabrera-Mireles et al., 2025) but in line with earlier work (Garcia-Guttierez & Gonzalez Maldonado, 2010). Equally, the 93-100% peak efficacy of *B. thuringiensis* against *P. xylostella* corresponds with published data (Wu et al., 2022; Zhou et al., 2025), whereas the 80-94% peak efficacy of granuloviruses is also accurate (Farrar et al., 2007). Peak predator-induced pest suppression in the field averaged 73.1 ± 5.6% (*B. tabaci*), 75.5 ± 5.7 (*H. armigera*) and 84.6 ± 6.3% (*P. xylostella*); data that align well with published averages of 73% across crops, pests, and geographies (Boldorini et al., 2024). LLM-generated estimates also confirm the potency of botanicals such as azadirachtin, pyrethrins or extracts from the Annonaceae (soursop family; Isman, 2020) or the role of agroecological preventative measures such as crop diversification (Harrison et al., 2019; Tamburini et al., 2020). Similarly, DeepSeek correctly synthesized the general impact profiles and predominant microbial, botanical or agroecological solutions for weed or disease management. Consequently, even though machine-driven literature synthesis may be prone to factual inaccuracies, data fabrication and omission, both bots regularly detected the predominant trends in non-chemical pest management.

    Today's generation of LLMs present shortcomings, which may restrict their utility in developing detailed evidence syntheses or in providing dependable decision-support (Bolaños et al., 2024; Scherbakov et al., 2025; Susnjak et al., 2025). Nevertheless, they do carry merit for synthesizing coarse-grained patterns and bringing scientific knowledge to the masses. While the less precise AI outputs may still hold value for farmers, agri-food value chain actors and policymakers, opportunities on a scientific front may be more limited. Even though the current generation of AI models carry less value for steering pest management science, one can possibly still delegate small, focused tasks in the academic workflow to LLMs. Web-grounded models in particular analyze and extract data from large volumes of digitized text including scientific documents in the public domain - reportedly covering up to 4,691 publications (i.e., for agroecological solutions against *H. armigera*) in one single query. This, together with its correct portrayal of coarse-grained efficacy patterns, may make DeepSeek and possibly the paid version of ChatGPT 4o or its plugin ScholarAI suitable to capture general trends e.g., as done in the medical domain (Van Veen et al., 2024). Though we are still some time removed from automated or unsupervised i.e., machine-only reviews, chatbots

can also reliably support literature searching and relevance screening (Spillias et al., 2024; Susnjak et al., 2025). In the interim, agricultural scientists undoubtedly already use chatbots for manuscript development (Kobak et al., 2025; Liang et al., 2025).

Looking ahead, joint human-machine efforts are likely the next step to harness the power of generative AI in fields such as agroecology or sustainable crop protection (Gurr et al., 2024; Spillias et al., 2024). Our results show that machine-only decision support carries shortcomings, which can only be picked up by expert verification (Agathokleous et al., 2023). As up to 70% of AI-generated references may be inaccurate and nonsensical output is often spun (Kacena et al., 2024), human oversight, rigorous fact-checking and robust evaluation frameworks are imperative (Van Dis et al., 2023; Bolaños et al., 2024; Schmidt et al., 2024). These limitations constrain their potential use in the scientific enterprise – where trustworthy data are vital. As such, when relying upon general-purpose LLMs, researchers, IT professionals and policy makers must adopt due vigilance, pursue transparency and accountability while guaranteeing data quality. These barriers may pose less of an issue to other stakeholders; even coarse-grained information on the efficacy of agroecological measures or biopesticides can inform farm practice. As such, AI tools could generate some of the "actionable knowledge" that farmers require to confidently transition away from pesticide-intensive agriculture (Geertsema et al., 2016; Chaplin-Kramer et al., 2019). Yet, issues may still arise when AI-generated outputs are integrated into proprietary decision support systems (DSS) – which can cause deskilling as farmers switch to 'auto-pilot'. In the agri-food sector which is deeply interwoven in human society, policymakers will need to ensure responsible use of AI-based technologies (De Clercq et al., 2024).

Our work shows that, regardless of its present shortfalls, machine power can mobilize critical knowledge and empower (non-scientist) actors in their pursuit of global sustainability ambitions (Candel, 2022; Schneider et al., 2024). By putting scientific knowledge into the hands of practitioners, policy makers and the general public, AI tools can lift critically under-funded scientific domains such as agroecology or biological control (van Lenteren, 2012; Pavageau et al., 2020) and push the boundaries of nature-friendly farming. Handled with prudency and care, web-grounded models in particular may distribute scientific knowledge more equitably and thereby act as a powerful catalyst for science-driven agroecological transitions.

## Acknowledgements

This work was funded by and executed by the United Nations Food and Agriculture Organization (FAO). We are grateful to Jeffery W. Bentley for revising an earlier draft of the manuscript.

## Author contributions

KAGW led the idea generation, data collection, interpretation and analysis, writing, and editing process.

## Data availability statement

All data underlying any of the analyses presented in this manuscript will be facilitated upon request.

## Literature references

Agathokleous, E., Saitanis, C.J., Fang, C. and Yu, Z. 2023. Use of ChatGPT: What does it mean for biology and environmental science?. Science of The Total Environment, 888, p.164154.

Alsahli, A.A., Alaraidh, I.A., Rashad, Y.M. and Razik, E.S.A. 2018. Extract from *Curcuma longa* L. triggers the sunflower immune system and induces defence-related genes against Fusarium root rot. Phytopathologia Mediterranea, 57(1), 26-36.

Anonymous, 2024. China has become a scientific superpower. https://www.economist.com/science-and-technology/2024/06/12/china-has-become-a-scientific-superpower, accessed on August 20, 2025.

Avanço, G.B., Ferreira, F.D., Bomfim, N.S., Peralta, R.M., Brugnari, T., Mallmann, C.A., de Abreu Filho, B.A., Mikcha, J.M.G. and Machinski Jr, M. 2017. *Curcuma longa* L. essential oil composition, antioxidant effect, and effect on *Fusarium verticillioides* and fumonisin production. Food Control, 73, 806-813.

Baidoo-Anu, D., & Ansah, L. O. 2023. Education in the era of generative artificial intelligence (AI): Understanding the potential benefits of ChatGPT in promoting teaching and learning. Journal of AI, 7(1), 52-62.

Bolaños, F., Salatino, A., Osborne, F., & Motta, E. 2024. Artificial intelligence for literature reviews: opportunities and challenges. Artificial Intelligence Review, 57(10), 259.

Boldorini, G.X., Mccary, M.A., Romero, G.Q., Mills, K.L., Sanders, N.J., Reich, P.B., Michalko, R. and Gonçalves-Souza, T. 2024. Predators control pests and increase yield across crop types and climates: a meta-analysis. Proceedings of the Royal Society B, 291(2018), p.20232522.

Bommasani, R., Hudson, D.A., Adeli, E., Altman, R., Arora, S., Von Arx, S., Bernstein, M.S., Bohg, J., Bosselut, A., Brunskill, E. and Brynjolfsson, E. 2021. On the opportunities and risks of foundation models. arXiv 2021. arXiv preprint arXiv:2108.07258.

Boyette, C.D. and Hoagland, R.E., 2015. Bioherbicidal potential of *Xanthomonas campestris* for controlling *Conyza canadensis*. Biocontrol Science and Technology, 25(2), 229-237.

Cabrera-Mireles, H., Jiménez-Jiménez, M., Ruiz-Ramírez, J., Murillo-Cuevas, F.D., Adame-García, J., Jiménez-Zilli, J., Vásquez Hernández, A. and Herrera-Bonilla, R.U. 2025. Microbial biopesticides to control whiteflies in eggplant *Solanum melongena*, in greenhouse. Insects, 16(6), p.578.

Calone, R., Raparelli, E., Bajocco, S., Rossi, E., Crecco, L., Morelli, D., Bassi, C., Tiso, R., Bugiani, R., Pietrangeli, F. and Cattaneo, G. 2025. Analysing the potential of ChatGPT to support plant disease risk forecasting systems. Smart Agricultural Technology, 10, p.100824.

Candel, J. 2022. EU food-system transition requires innovative policy analysis methods. Nature Food, 3(5), 296-298.

Capraro, V., Lentsch, A., Acemoglu, D., Akgun, S., Akhmedova, A., Bilancini, E., Bonnefon, J.F., Brañas-Garza, P., Butera, L., Douglas, K.M. et al. 2024. The impact of generative artificial intelligence on socioeconomic inequalities and policy making. PNAS Nexus, 3(6), p.pgae191.

Chaplin-Kramer, R., O'Rourke, M., Schellhorn, N., Zhang, W., Robinson, B.E., Gratton, C., Rosenheim, J.A., Tscharntke, T. and Karp, D.S., 2019. Measuring what matters: actionable information for conservation biocontrol in multifunctional landscapes. Frontiers in Sustainable Food Systems, 3, p.60.

Chen, H., Jiang, Z., Liu, X., Xue, C. C., Yew, S. M. E., Sheng, B. et al. 2025. Can large language models fully automate or partially assist paper selection in systematic reviews?. British Journal of Ophthalmology, in press.

Conroy, G. and Mallapaty, S. 2025. How China created AI model DeepSeek and shocked the world. Nature, 638, 300-301.

De Clercq, D., Nehring, E., Mayne, H. and Mahdi, A. 2024. Large language models can help boost food production, but be mindful of their risks. Frontiers in Artificial Intelligence, 7, p.1326153.

Deguine, J.P., Aubertot, J.N., Bellon, S., Côte, F., Lauri, P.E., Lescourret, F., Ratnadass, A., Scopel, E., Andrieu, N., Bàrberi, P. and Becker, N. 2023. Agroecological crop protection for sustainable agriculture. Advances in Agronomy, 178, 1-59.

Emsley, R. 2023. ChatGPT: these are not hallucinations–they're fabrications and falsifications. Schizophrenia, 9(1), p.52.

Faria, M. and Wraight, S.P. 2001. Biological control of *Bemisia tabaci* with fungi. Crop Protection, 20(9), 767-778.

Farmer, E.E., Brown, D., Gore, M.A. and Tufan, H.A. 2025. Applying large language models to extract information from crop trait prioritization studies. Plants, People, Planet, in press.

Farrar, R.R., Shapiro, M. and Shepard, M. 2007. Relative activity of baculoviruses of the diamondback moth, *Plutella xylostella* (L.)(Lepidoptera: Plutellidae). BioControl, 52(5), 657-667.

Furlong, M.J., Wright, D.J. and Dosdall, L.M. 2013. Diamondback moth ecology and management: problems, progress, and prospects. Annual Review of Entomology, 58(1), 517-541.

Gibney, E. 2025. China's cheap, open AI model DeepSeek thrills scientists. Nature, 638, 13-14.

Gartlehner G, Affengruber L, Titscher V, et al. 2020. Single-reviewer abstract screening missed 13 percent of relevant studies: a crowdbased, randomized controlled trial. Journal of Clinical Epidemiology, 121, 20-28.

Geertsema, W., Rossing, W.A., Landis, D.A., Bianchi, F.J., Van Rijn, P.C., Schaminée, J.H., Tscharntke, T. and Van Der Werf, W. 2016. Actionable knowledge for ecological intensification of agriculture. Frontiers in Ecology and the Environment, 14(4), 209-216.

Gohr, C., Rodríguez, G., Belomestnykh, S., Berg-Moelleken, D., Chauhan, N., Engler, J.O., Heydebreck, L.V., Hintz, M.J., Kretschmer, M., Krügermeier, C. et al., 2025. Artificial intelligence in sustainable development research. Nature Sustainability, 8, 970–978.

Gousi, F., Belabess, Z., Laboureau, N., Peterschmitt, M. and Pooggin, M.M. 2025. A novel parvovirus associated with the whitefly *Bemisia tabaci*. Pathogens, 14(7), 714.

Gurr, G.M., Liu, J. and Pogrebna, G. 2024. Harnessing artificial intelligence for analysing the impacts of nectar and pollen feeding in conservation biological control. Current Opinion in Insect Science, 62, p.101176.

Harrison, R.D., Thierfelder, C., Baudron, F., Chinwada, P., Midega, C., Schaffner, U. and Van Den Berg, J. 2019. Agro-ecological options for fall armyworm (*Spodoptera frugiperda* JE Smith) management: providing low-cost, smallholder friendly solutions to an invasive pest. Journal of Environmental Management, 243, 318-330.

Hoagland, R.E., Boyette, C.D. and Stetina, K.C. 2023. Bioherbicidal activity of *Albifimbria verrucaria* (formerly *Myrothecium verrucaria*) on glyphosate-resistant *Conyza Canadensis*. Journal of Fungi 9(7), p.773.

Hosseini, R., Schmidt, O. and Keller, M.A. 2012. Detection of predators within Brassica crops: a search for predators of diamondback moth (*Plutella xylostella*) and other important pests. African Journal of Agricultural Research 7(23), 3473-3484.

Isman, M.B. 2020. Botanical insecticides in the twenty-first century—fulfilling their promise?. Annual Review of Entomology 65(1), 233-249.

Janssen, A. and van Rijn, P.C. 2021. Pesticides do not significantly reduce arthropod pest densities in the presence of natural enemies. Ecology letters 24(9), 2010-2024.

Jin, I., Tangsrivimol, J.A., Darzi, E., Hassan Virk, H.U., Wang, Z., Egger, J., Hacking, S., Glicksberg, B.S., Strauss, M. and Krittanawong, C. 2025. DeepSeek vs. ChatGPT: prospects and challenges. Frontiers in Artificial Intelligence 8, p.1576992.

Kacena, M.A., Plotkin, L.I., & Fehrenbacher, J.C. 2024. The use of artificial intelligence in writing scientific review articles. Current Osteoporosis Reports, 22(1), 115-121.

Kaygisiz, Ö.F. and Teke, M.T. 2025. Can deepseek and ChatGPT be used in the diagnosis of oral pathologies?. BMC Oral Health, 25(1), p.638.

Khraisha, Q., Put, S., Kappenberg, J., Warraitch, A. and Hadfield, K. 2024. Can large language models replace humans in systematic reviews? Evaluating GPT-4's efficacy in screening and extracting data from peer-reviewed and grey literature in multiple languages. Research Synthesis Methods, 15(4), 616-626.

Kim, M.K., Choi, G.J. and Lee, H.S., 2003. Fungicidal property of *Curcuma longa* L. rhizome-derived curcumin against phytopathogenic fungi in a greenhouse. Journal of Agricultural and Food Chemistry, 51(6), 1578-1581.

Kim, M.J., Shim, C.K. and Park, J.H. 2021. Control efficacy of *Bacillus velezensis* AFB2-2 against potato late blight caused by *Phytophthora infestans* in organic potato cultivation. The Plant Pathology Journal, 37(6), p.580.

Kobak, D., González-Márquez, R., Horvát, E.Á. and Lause, J. 2025. Delving into LLM-assisted writing in biomedical publications through excess vocabulary. Science Advances, 11(27), p.eadt3813.

Kung, T.H., Cheatham, M., Medenilla, A., Sillos, C., De Leon, L., Elepaño, C., Madriaga, M., Aggabao, R., Diaz-Candido, G., Maningo, J. and Tseng, V. 2023. Performance of ChatGPT on USMLE: potential for AI-assisted medical education using large language models. PLoS Digital Health, 2(2), p.e0000198.

Liang, W., Zhang, Y., Wu, Z., Lepp, H., Ji, W., Zhao, X., Cao, H., Liu, S., He, S., Huang, Z. and Yang, D. 2025. Quantifying large language model usage in scientific papers. Nature Human Behaviour, pp.1-11.

Luo, P.W., Liu, J.W., Xie, X., Jiang, J.W., Huo, X.Y., Chen, Z.L., Huang, Z.C., Jiang, S.Q. and Li, M.Q. 2025. DeepSeek vs ChatGPT: A comparison study of their performance in answering prostate cancer radiotherapy questions in multiple languages. American Journal of Clinical and Experimental Urology, 13(2), p.176.

Marston, J. 2023. Why farmers business network launched Norm, an AI Advisor for farmers built on ChatGPT. San Francisco, CA: AgFunderNews.

Mollah, M.M.I. and Hassan, N. 2023. Efficacy of *Trichoderma harzianum*, as a biological fungicide against fungal diseases of potato, late blight and early blight. Journal of Natural Pesticide Research, 5, p.100047.

Mostafapour, M., Fortier, J.H., Pacheco, K., Murray, H. and Garber, G. 2024. Evaluating literature reviews conducted by humans versus ChatGPT: comparative study. Jmir ai, 3, p.e56537.

Normile, D. 2025. Chinese firm's large language model makes a splash. Science 387, 238.

Oerke, E.C., 2006. Crop losses to pests. The Journal of Agricultural Science, 144(1), 31-43.

OpenAI, 2024 Hello GPT-4o. Available online at: https://openai.com/index/hello-gpt-4o/ (Accessed August 15, 2025).

Pavageau, C., Pondini, S. and Geck, M. 2020. Money flows: what is holding back investment in agroecological research for Africa. Biovision Foundation for Ecological Development & International Panel of Experts on Sustainable Food Systems IPES-Food, Zurich, Switzerland.

Perrot, T., Möhring, N., Rusch, A., Gaba, S. and Bretagnolle, V. 2025. Crop yield loss under high insecticide regime driven by reduction in natural pest control. Proceedings B, 292(2051), p.20250138.

Quan, X., Wu, L., Zhou, Q., Yun, Y., Peng, Y., & Chen, J. 2011. Identification of predation by spiders on the diamondback moth *Plutella xylostella*. Bulletin of Insectology, 64(2), 223-227.

Schaffner, U., Heimpel, G.E., Mills, N.J., Muriithi, B.W., Thomas, M.B. and Wyckhuys, K.A.G. 2024. Biological control for One Health. Science of The Total Environment, 951, p.175800.

Scherbakov, D., Hubig, N., Jansari, V., Bakumenko, A. and Lenert, L.A. 2025. The emergence of large language models as tools in literature reviews: a large language model-assisted systematic review. Journal of the American Medical Informatics Association, 32(6), 1071-1086.

Schmidt, L., Hair, K., Graziosi, S., Campbell, F., Kapp, C., Khanteymoori, A. et al. 2024. Exploring the use of a large language model for data extraction in systematic reviews: a rapid feasibility study. arXiv preprint arXiv:2405.14445.

Schneider, K., Barreiro-Hurle, J. and Rodriguez-Cerezo, E. 2023. Pesticide reduction amidst food and feed security concerns in Europe. Nature Food, 4(9), 746-750.

Shattuck, A., Werner, M., Mempel, F., Dunivin, Z. and Galt, R. 2023. Global pesticide use and trade database (GloPUT): New estimates show pesticide use trends in low-income countries substantially underestimated. Global Environmental Change, 81, p.102693.

Shepherd, K.D., Miller, M.A., Kisitu, B., Miles, B.G., Gbedevi, K., Chunga, P.Z., Thuo, A.G.W., Musiiwa, R.F., Gyampo, E.D., Cross, M. et al., 2025. Virtual Agronomist-an AI-assisted chatbot for guiding crop management decisions of smallholder farmers in Africa. agriRxiv, (2025), p.20250269995.

Silva, B., Nunes, L., Esteva, R., Aski, V., and Chandra, R. 2023. GPT-4 as an agronomist assistant? Answering agriculture exams using large language models. arXiv [preprint] arXiv.2310.06225. doi: 10.48550/arXiv.2310.06225

Spillias, S., Tuohy, P., Andreotta, M., Annand-Jones, R., Boschetti, F., Cvitanovic, C. et al. 2024. Human-AI collaboration to identify literature for evidence synthesis. Cell Reports Sustainability, 1(7).

Struelens, Q.F., Rivera, M., Alem Zabalaga, M., Ccanto, R., Quispe Tarqui, R., Mina, D., Carpio, C., Yumbla Mantilla, M.R., Osorio, M., Roman, S. et al. 2022. Pesticide misuse among small Andean farmers stems from pervasive misinformation by retailers. PLOS Sustainability and Transformation, 1(6), p.e0000017.

Suraporn, S. and Terenius, O. 2021. Supplementation of *Lactobacillus casei* reduces the mortality of Bombyx mori larvae challenged by *Nosema bombycis*. BMC Research Notes, 14(1), p.398.

Susnjak, T., Hwang, P., Reyes, N., Barczak, A. L., McIntosh, T., & Ranathunga, S. 2025. Automating research synthesis with domain-specific large language model fine-tuning. ACM Transactions on Knowledge Discovery from Data, 19(3), 1-39.

Tamburini, G., Bommarco, R., Wanger, T.C., Kremen, C., Van Der Heijden, M.G., Liebman, M. and Hallin, S. 2020. Agricultural diversification promotes multiple ecosystem services without compromising yield. Science Advances, 6(45), p.eaba1715.

Tang, F.H., Wyckhuys, K.A., Li, Z., Maggi, F. and Silva, V. 2025. Transboundary impacts of pesticide use in food production. Nature Reviews Earth & Environment, 6, 383–400.

Tollefson, J. 2025. China's scientific clout is growing as US influence wanes: the data show howhttps://www.nature.com/articles/d41586-025-03956-y, accessed on December 12, 2025.

Tordjman, M., Liu, Z., Yuce, M., Fauveau, V., Mei, Y., Hadjadj, J., Bolger, I., Almansour, H., Horst, C., Parihar, A.S. and Geahchan, A. 2025. Comparative benchmarking of the DeepSeek large language model on medical tasks and clinical reasoning. Nature Medicine, pp.1-1.

Van Dis, E.A., Bollen, J., Zuidema, W., Van Rooij, R. and Bockting, C.L. 2023. ChatGPT: five priorities for research. Nature, 614(7947), 224-226.

Van Lenteren, J.C. 2012. The state of commercial augmentative biological control: plenty of natural enemies, but a frustrating lack of uptake. BioControl, 57(1), 1-20.

Van Veen, D., Van Uden, C., Blankemeier, L., Delbrouck, J.B., Aali, A., Bluethgen, C., Pareek, A., Polacin, M., Reis, E.P., Seehofnerová, A. and Rohatgi, N. 2024. Adapted large language models can outperform medical experts in clinical text summarization. Nature Medicine, 30(4), 1134-1142.

Vidal, C., Lace, L.A. and Fargues, J. 1997. Pathogenicity of *Paecilomyces fumosoroseus* (Deuteromycotina: Hyphomycetes) against *Bemisia argentifolii* (Homoptera: Aleyrodidae) with a description of a bioassay method. Journal of Economic Entomology, 90(3), 765-772.

Wu, L.H., Chen, Y.Z., Hsieh, F.C., Lai, C.T. and Hsieh, C. 2022. Combined effect of *Photorhabdus luminescens* and *Bacillus thuringiensis* subsp. *aizawai* on *Plutella xylostella*. Applied Microbiology and Biotechnology, 106(8), 2917-2926.

Yang, S., Yuan, Z., Li, S., Peng, R., Liu, K. and Yang, P. 2024. Gpt-4 as evaluator: Evaluating large language models on pest management in agriculture. arXiv preprint arXiv:2403.11858.

Yuan XT, Shao CY, Zhang ZZ and Qian D. 2025. Comparing the performance of ChatGPT and ERNIE Bot in answering questions regarding liver cancer interventional radiology in Chinese and English contexts: a comparative study. Digital Health 11, 20552076251315511.

Zhou, L., Wang, Y., Liu, Z.Y., Liu, X., Zhai, Z., Cao, S.K., Zhao, Q.Q., Zaghloul, H.A., Shi, X.B., Yu, H. and Su, H. 2025. Biological activity and field efficacy of *Bacillus thuringiensis kurstaki* strains with protein film adjuvants (PFAs) against *Plutella xylostella* (Lepidoptera: Plutellidae). Pest Management Science, 81(7), 4108-4120.

Zhou, S., Zhang, J., Lin, Y., Li, X., Liu, M., Hafeez, M., Huang, J., Zhang, Z., Chen, L., Ren, X. and Dong, W. 2023. *Spodoptera exigua* multiple nucleopolyhedrovirus increases the susceptibility to insecticides: a promising efficient way for pest resistance management. Biology, 12(2), p.260.

*Bemisia tabaci*

A
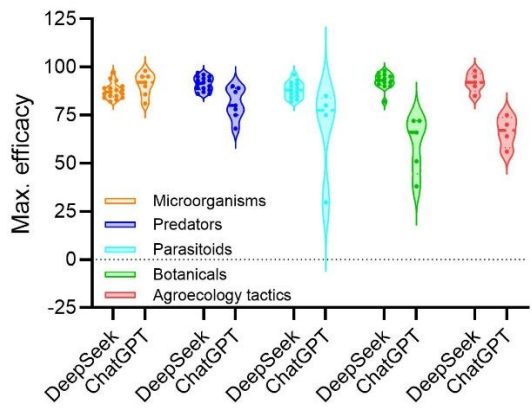

B
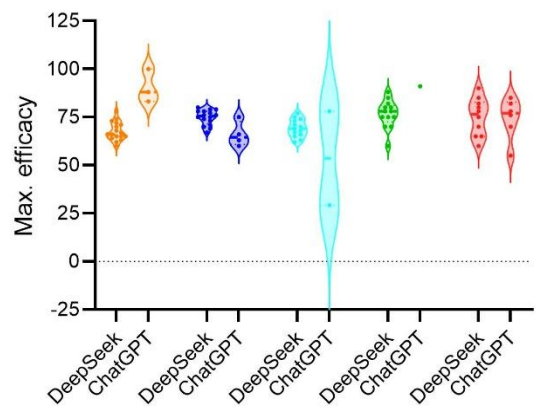

*Helicoverpa armigera*

A
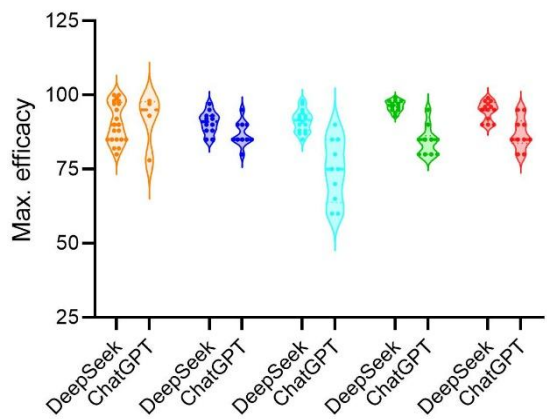

B
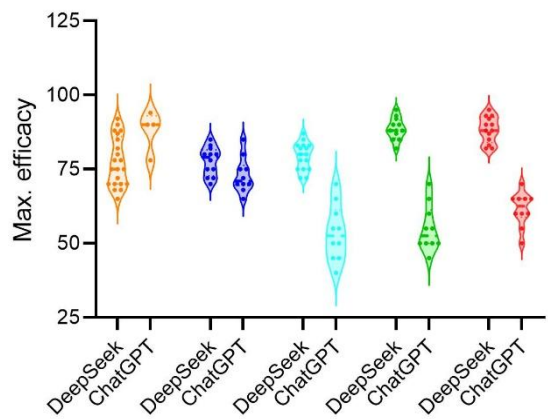

*Plutella xylostella*

A
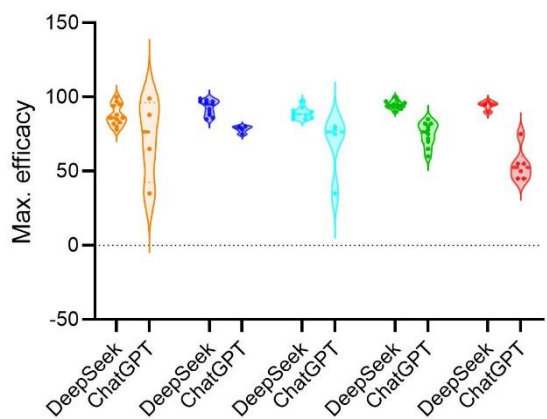

B
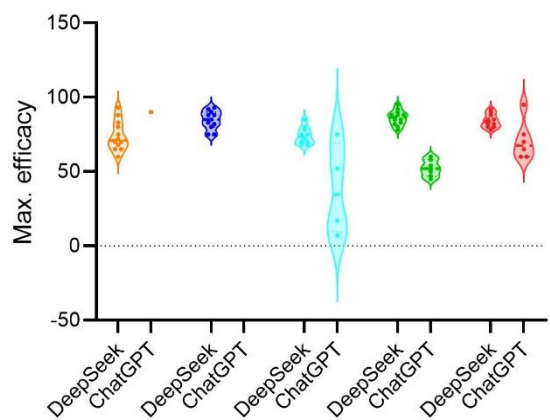

**Figure 1**. **Maximum laboratory (A) and field (B) efficacy of non-chemical management strategies for three insect pests, as reported by two AI engines**. Patterns are plotted for microbial or invertebrate biological control agents, botanical insecticides and agroecological measures, tested against *B. tabaci*, *H. armigera* or *P. xylostella*. All data are AI-generated and thus potentially fictitious.

DeepSeek
A

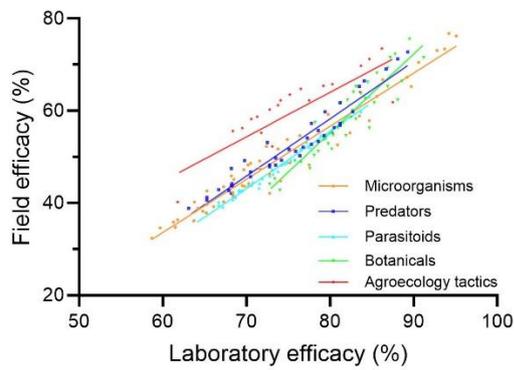

B

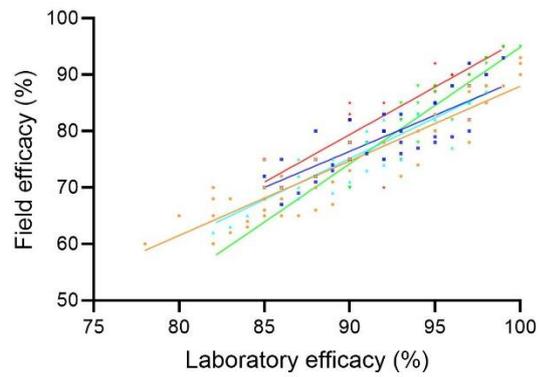

ChatGPT
A

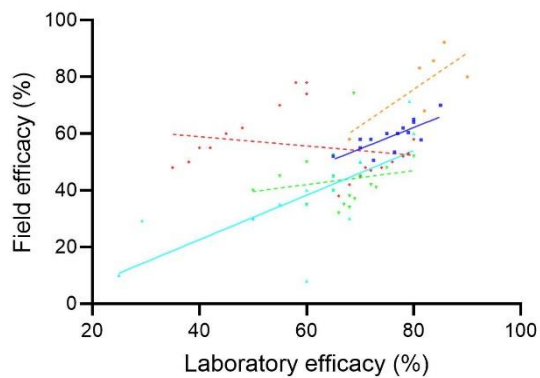

B

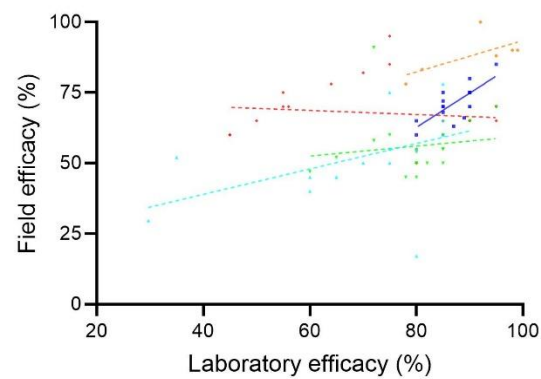

**Figure 2. Agreement between AI-reported field and laboratory efficacy data for five non-chemical pest management strategies.** Patterns are plotted for average (A) and maximum (B) efficacy records, as reported by either DeepSeek or ChatCGPT. Data are consolidated for three globally important pests i.e., *B. tabaci*, *H. armigera* and *P. xylostella*. Linear regression lines are drawn for each individual management strategy, with dashed lines representing non-significant patterns (ANOVA, p> 0.05).

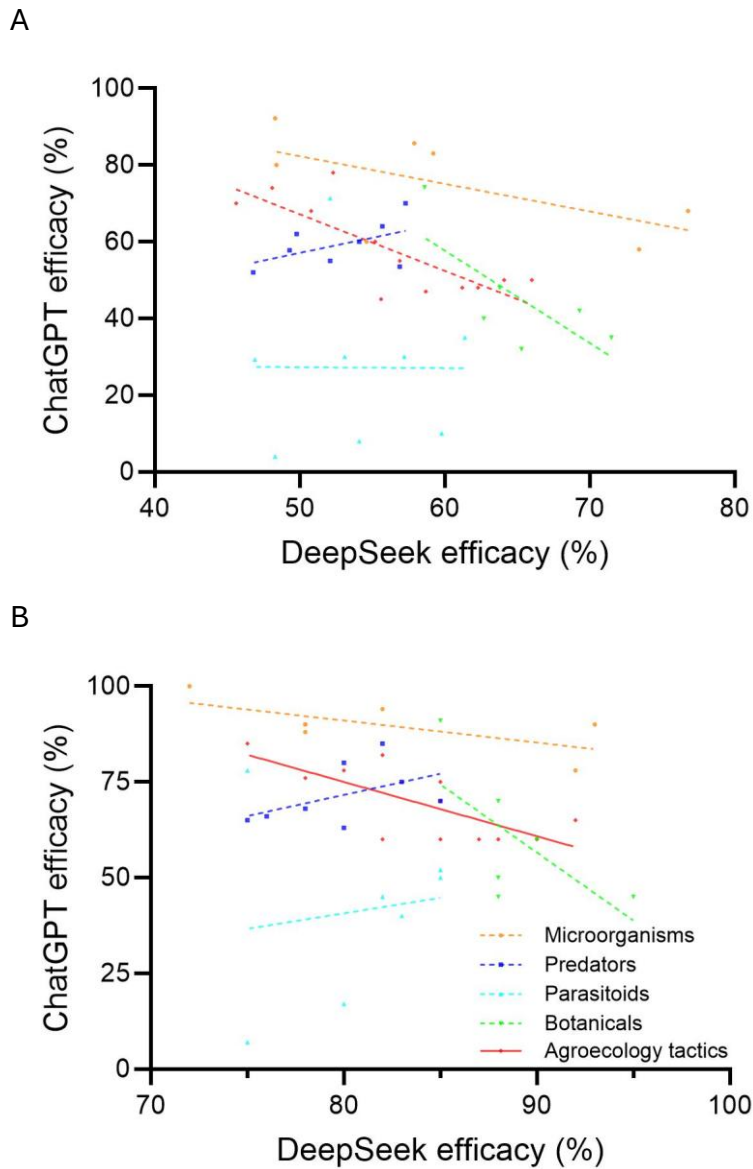

**Figure 3. Agreement between pest management efficacy data reported by either DeepSeek or ChatCGPT.** Pest management efficacy data are consolidated for three key pests i.e., *B. tabaci*, *H. armigera* and *P. xylostella*. Patterns are plotted for each of five categories of non-chemical pest management, with each data point representing one given management solution, compound or agent. Trendlines represent regression results for AI-reported average (A) and maximum (B) field efficacy levels, with dashed lines representing non-significant patterns (ANOVA, p>0.05).

**Table 1. Breadth of literature screening by DeepSeek or ChatGPT, when documenting the efficacy of non-chemical *Bemisia tabaci* management tactics in China.** Disparity between the number of literature records covered at each given step of the PRISMA flow diagram is expressed as relative or percentage difference (%). Negative values reflect instances in which higher values were obtained through ChatGPT than DeepSeek.

|  | Data disparity or database coverage | | | | |
|---|---|---|---|---|---|
|  | **Microbials** | **Predators** | **Parasitoids** | **Botanicals** | **Agroecology** |
| **PRISMA flow diagram** | | | | | |
| Initial records identified | 147% (724)[a] | 12% (1103) | 17% (512) | 146% (638) | 122% (1210) |
| Duplicates removed | 169% (161) | -30% (121) | 123% (121) | 163% (147) | 143% (290) |
| Records screened | 142% (563) | 19% (982) | -3% (391) | 141% (491) | 116% (920) |
| Excluded at title/abstract level | 124% (297) | -105% (218) | -42% (203) | 123% (253) | 111% (650) |
| Full-texts assessed | 166% (266) | 149% (764) | 70% (188) | 162% (238) | 129% (270) |
| Excluded with reasons | 149% (89) | 173% (632) | -21% (63) | 163% (87) | 111% (177) |
| Studies included | 175% (177) | 68% (132) | 162% (125) | 162% (151) | 172% (93) |
| **Literature resources screened** | | | | | |
| ISI-indexed databases | | | | | |
|   Web of Science | 1/1[b] | 1/1 | 1/1 | 1/1 | 1/1 |
|   Scopus | 1/1 | 1/1 | 1/1 | 1/1 | 1/1 |
| Chinese databases | | | | | |
|   CNKI[c] | 1/1 | 1/1 | 1/1 | 1/1 | 1/1 |
|   Wanfang | 1/0 | 1/1 | 1/1 | 1/0 | 1/0 |
|   CQVIP | 1/0 | 1/1 | 1/1 | 1/0 | 1/0 |
|   CSTJ | 0/0 | 0/0 | 0/1 | 0/0 | 0/0 |
| Specialized databases | | | | | |
|   CAB Abstracts | 1/1 | 1/1 | 1/1 | 1/1 | 1/1 |
|   PubAG (USA) | 0/0 | 0/1 | 0/0 | 0/0 | 0/0 |
|   AGRICOLA / AGRIS (FAO) | 1/1 | 1/1 | 1/1 | 1/1 | 1/1 |
| National repositories | | | | | |
|   Institutional repositories | 1/1 | 1/0 | 1/0 | 1/1 | 0/0 |
|   Provincial agricultural bulletins | 1/0 | 1/0 | 1/0 | 1/0 | 0/0 |
| Others e.g., SpringerLink | 0/0 | 0/1 | 0/1 | 0/0 | 0/0 |

a. Number of literature records reported by DeepSeek
b. Reported coverage of a respective database by DeepSeek (numerator) or ChatGPT (denominator)
c. China National Knowledge Infrastructure (CNKI), Chongqing VIP (CQVIP), China Science and Technology Journal Database (CSTJ)

**Table 2**. **Laboratory and field efficacy for non-chemical disease management tactics in China and globally, as reported by DeepSeek.** Data are shown for three globally limiting plant diseases i.e., potato late blight, wheat rust and Fusarium head blight. For each disease and tactic, the three best-performing agents or solutions are listed. All data and listed agents or solutions are AI-generated and may thus be fictitious.

| Scope | Scientific base[1] | Max. lab efficacy | Max. field efficacy | Top-3 solutions (in-field) |
|---|---|---|---|---|
| **Potato late blight** | | | | |
| *Microorganisms* | | | | |
| Global | 27.4 ± 11.6 | 94.6 ± 3.3 | 76.9 ± 7.7 (88%)[2] | Mixtures (88%); *Pseudomonas fluorescens* (82%); *Paenibacillus polymyxa* (80%) |
| China | 10.5 ± 4.8 | 92.3 ± 3.9 | 73.7 ± 8.4 (85%) | |
| *Botanicals* | | | | |
| Global | 37.9 ± 10.2 | 95.7 ± 2.6 | 82.6 ± 5.4 (90%) | Mixtures (90%); turmeric extract (88%); garlic extract (85%) |
| China | 12.3 ± 3.6 | 93.1 ± 3.5 | 77.3 ± 8.1 (88%) | |
| *Agroecology measures* | | | | |
| Global | 82.6 ± 19.4 | 89.6 ± 5.9 | 76.4 ± 9.1 (90%) | Multi-methods (90%); Si amendments (85%); bio-fumigants (80%) |
| China | 24.4 ± 6.8 | 87.9 ± 5.0 | 71.9 ± 8.7 (85%) | |
| **Wheat rust** | | | | |
| *Microorganisms* | | | | |
| Global | 40.0 ± 15.1 | 93.4 ± 3.9 | 74.1 ± 7.6 (85%) | Mixtures (85%); *Pseudomonas fluorescens* (80%); *Paenibacillus polymyxa* (77%) |
| China | 13.0 ± 5.2 | 91.6 ± 4.0 | 72.1 ± 8.3 (85%) | |
| *Botanicals* | | | | |
| Global | 51.3 ± 16.1 | 95.7 ± 2.6 | 83.3 ± 6.2 (92%) | Mixtures (92%); turmeric extract (90%); garlic extract (85%) |
| China | 14.9 ± 5.1 | 93.3 ± 3.5 | 78.3 ± 6.6 (88%) | |
| *Agroecology measures* | | | | |
| Global | 92.2 ± 24.7 | 91.2 ± 5.8 | 77.5 ± 9.4 (90%) | Multi-methods (90%); Si amendments (85%); biofumigants (80%) |
| China | 21.2 ± 5.5 | 87.2 ± 7.4 | 73.3 10.2 (88%) | |
| **Fusarium head blight** | | | | |
| *Microorganisms* | | | | |
| Global | 55.8 ± 20.1 | 92.5 ± 3.3 | 74.6 ± 6.3 (85%) | Mixtures (85%); *Pseudomonas fluorescens* (80%); *Clonostachys rosea* (78%) |
| China | 17.3 ± 7.0 | 89.7 ± 3.5 | 70.0 ± 6.6 (80%) | |
| *Botanicals* | | | | |
| Global | 88.1 ± 25.6 | 96.4 ± 1.7 | 86.1 ± 5.0 (93%) | Mixtures (93%); turmeric extract (92%); garlic extract (88%) |
| China | 17.3 ± 5.5 | 94.3 ± 2.9 | 79.3 ± 7.7 (90%) | |
| *Agroecology measures* | | | | |
| Global | 105.0 ± 24.3 | 92.5 ± 4.7 | 81.7 ± 8.3 (92%) | Multi-methods (92%); Si amendments (88%); biofumigants (80%) |
| China | 22.7 ± 6.2 | 90.7 ± 4.5 | 78.3 ± 8.3 (90%) | |

[1]. Number of publications per organism or solution
[2]. Highest efficacy level reported between brackets

**Table 3**. Laboratory and field efficacy for non-chemical weed management tactics in China and globally, as reported by DeepSeek. Data are shown for three globally limiting agricultural weeds i.e., *Echinochloa* spp., *Erigeron* (or *Conyza*) *canadensis* and *Amaranthus* spp. For each disease and tactic, the three best-performing agents or solutions are listed. All data and listed agents or solutions are AI-generated and may thus be fictitious.

| Scope | Scientific base[1] | Max. lab efficacy | Max. field efficacy | Top-3 solutions globally (in-field) |
|---|---|---|---|---|
| *Echinochloa* spp. | | | | |
| Microorganisms | | | | |
| Global | 37.2 ± 11.6 | 95.8 ± 2.3 | 86.3 5.3 (95%)[2] | Mixtures (95%); *Drechslera gigantea* (88%); *Exserohilum monoceras* (88%) |
| China | 12.5 ± 4.0 | 94.8 ± 3.3 | 80.0 6.6 (90%) | |
| Botanicals | | | | |
| Global | 85.2 ± 20.8 | 94.6 ± 3.0 | 83.4 6.7 (92%) | Mixtures (92%); saponin-rich extracts (88%); essential oils (80%) |
| China | 18.7 ± 4.7 | 91.2 ± 4.4 | 78.3 7.3 (88%) | |
| Agroecology measures | | | | |
| Global | 41.7 ± 16.1 | 94.1 ± 5.0 | 84.0 ± 5.0 (90%) | Allelopathic intercrops (90%); mulching (88%); water management (85%) |
| China | 12.6 ± 5.4 | 93.9 ± 5.3 | 84.0 ± 5.0 (90%) | |
| *Erigeron* (*Conyza*) *canadensis* | | | | |
| Microorganisms | | | | |
| Global | 33.8 ± 14.4 | 94.5 ± 3.0 | 84.7 ± 6.2 (95%) | Mixtures (95%); *Alternaria alternata* (88%); *Fusarium proliferatum* (85%) |
| China | 10.6 ± 2.7 | 92.6 ± 3.6 | 77.6 ± 6.8 (88%) | |
| Botanicals | | | | |
| Global | 66.0 ± 16.4 | 95.2 ± 3.4 | 85.4 ± 7.6 (95%) | Mixtures (92%); saponin-rich extracts (90%); essential oils (85%) |
| China | 13.2 ± 3.5 | 92.7 ± 3.3 | 78.3 ± 6.3 (88%) | |
| Agroecology measures | | | | |
| Global | 31.7 ± 13.7 | 92.9 ± 4.8 | 83.6 ± 5.1 (90%) | Allelopathic intercrops (90%); mulching (88%); cover crops (85%) |
| China | 8.6 ± 4.0 | 88.7 ± 5.0 | 79.3 ± 5.3 (85%) | |
| *Amaranthus* spp. | | | | |
| Microorganisms | | | | |
| Global | 37.0 ± 16.4 | 94.0 ± 2.9 | 83.3 4.9 (92%) | Mixtures (92%); *Alternaria alternata* (85%); *Myrothecium verrucaria* (83%) |
| China | 9.0 ± 3.2 | 92.6 ± 3.6 | 78.2 6.6 (88%) | |
| Botanicals | | | | |
| Global | 91.4 ± 21.0 | 95.4 ± 3.4 | 85.6 ± 7.6 (95%) | Mixtures (95%); saponin-rich extracts (90%); essential oils (86%) |
| China | 16.7 ± 3.7 | 92.7 ± 3.3 | 78.3 ± 6.3 (88%) | |
| Agroecology measures | | | | |
| Global | 64.6 ± 30.4 | 95.4 ± 3.1 | 88.0 ± 4.4 (93%) | Allelopathic intercrops (93%); mulching (92%); water management (90%) |
| China | 17.9 ± 9.2 | 93.1 ± 4.4 | 84.0 ± 5.0 (90%) | |

[1]. Number of publications per organism or solution
[2]. Highest efficacy level reported between brackets

**Supplementary Table 1. Standardized AI queries or instruction prompts that were submitted to either of the two search engines.** A first query, targeting entomopathogenic microorganisms of *Bemisia tabaci* in China, also constituted the basis for queries that covered invertebrate biological control agents i.e., predators or parasitoids and botanicals. This was achieved by slightly adapting the below query e.g., changing 'product efficacy' to 'predation efficacy' or 'parasitism levels'. A second query solely addressed agroecological measures that were tested against *B. tabaci* under laboratory, greenhouse or screenhouse, and field conditions within China. Both queries were slightly modified to cover other pests, weeds and plant diseases in other countries or at the global level.

| Topic | AI query |
|---|---|
| *Bemisia tabaci* x microbial insecticides | Please list the total number of all-time, ISI peer-reviewed scientific publications on all microbial insecticides against Bemisia tabaci in China in a table, indicating the total number of publications, efficacy in the laboratory (average + range), efficacy in the green- or screenhouse (average + range) and efficacy in the field (average + range) for each type of microbial insecticide. Break down publication output and efficacy data by species of microorganism, regardless of strain. Express product efficacy as the % reduction in pest density post-treatment, corrected with Abbott's formula when the control mortality exceeded 10%. Add the standard deviation to the average values. Include publications in any language. Adopt the following exclusion or inclusion or exclusion criteria:<br><br>Inclusion criteria:<br>1. Publication Type: Only ISI peer-reviewed journal articles indexed in Core ISI-Indexed Databases, specialized agricultural databases or regional / national repositories and national institutional sources.<br>Timeframe: All-time through June 2025. No minimum impact factor for source journals.<br><br>2. Geographic Scope: Studies must explicitly report experiments conducted in the focal country (lab, greenhouse, or field). Microbial strains, parasitoids or predators tested must be isolated from the focal country or approved for use in national agriculture.<br><br>3. Target Pest and natural enemies: Focused solely on the focal pest (any biotype). Studies on mixed pest systems are included only if data from the target pest are separable. Parasitoids or predators must be either native to the focal country or reported to be present.<br><br>4. Efficacy Metrics: Required quantitative predation, parasitism or efficacy data: % reduction in whitefly populations (nymphs/adults).<br>Parasitism rate (%) = (Parasitized nymphs/Total exposed nymphs)×100<br>Required quantitative mortality data (e.g., percent parasitism, corrected mortality using Abbott's formula). Both mean and variability measures (SD or SEM convertible to SD). Minimum 3 biological replicates.<br><br>5. Experimental Design: Lab studies: Controlled conditions (temperature, RH, photoperiod specified). Greenhouse/screenhouse: Semi-controlled with host plants. Field trials: Must include untreated controls and replicate plots. Statistical robustness: Studies with ≥3 replicates, reported sample sizes and clear statistical methods (SD/SEM, ANOVA).<br><br>Exclusion Criteria: |
| | 1. Non-ISI or Non-Peer-Reviewed Sources: Excluded: Conference abstracts, theses, book chapters, or non-indexed journals.

2. Geographic Irrelevance: Studies conducted outside the focal country or with non-native strains (unless validated in-country).

4. Incomplete Data: Papers lacking raw mortality rates, sample sizes, or statistical variability measures (SD/SEM). Studies reporting only $LC_{50}$/$LD_{50}$ without field-relevant efficacy percentages. Studies reporting only ""presence/absence"" without efficacy quantification.

4. Non-Target Effects or Formulation Studies: Excluded if focused solely on non-target toxicity, genomics, or formulation chemistry without efficacy data. Excluded if focused on non-predatory or non-parasitism interactions (e.g., pollination). Molecular studies without efficacy data (e.g., gut content analysis alone).

5. Duplicate Data: Multiple papers from the same research group with overlapping datasets are counted " |
| *Bemisia tabaci* x agro-ecological measures | Please list the total number of all-time, ISI peer-reviewed scientific publications on all agroecological preventative measures against *Bemisia tabaci* (as target pest) in China in a table, indicating the total number of publications, efficacy in the laboratory (average + range), efficacy in the green- or screenhouse (average + range) and efficacy in the field (average + range) for each type of measure. Break down publication output and efficacy data by measure. Express efficacy as the % reduction in pest density post-treatment and correct this with Abbott's formula when the control mortality exceeded 10%. Add the standard deviation to the average values. Include publications in any language. Adopt the following exclusion or inclusion criteria:

Inclusion Criteria:
1. Publication Type: Included: Peer-reviewed articles indexed in ISI Web of Science, Scopus, or Core national databases. Specialized agricultural databases (e.g., CAB Abstracts, AGRIS). Articles in any language (English, Chinese, etc.) with translatable methods/results. Excluded: Conference abstracts, theses, book chapters, non-peer-reviewed reports. Journals not indexed in the above databases.

2. Geographic Scope: Included: Studies conducted exclusively in the focal country (lab, greenhouse, or field). Measures using locally sourced materials (e.g., local marigold varieties) or approved practices. Excluded: Studies outside the focal country or with non-validated imported technologies.

3. Target Pest and Measures: Included: Focused solely on the target pest (any biotype), excluding any whitefly different from Bemisia tabaci. Mixed-pest studies only if the target pest data are separable.
Agroecological measures (e.g., intercropping, mulches, biocontrol) with preventative (not just curative) action. Excluded: Synthetic insecticides, genetic modification, or non-agroecological methods.

4. Efficacy Metrics: Included: % reduction in target pest density (nymphs/adults) vs. untreated controls. Reported mean ± SD/SEM (SEM converted to SD if needed). Minimum 3 biological replicates. Excluded: Studies reporting only $LC_{50}$/$LD_{50}$ without field-relevant efficacy. Qualitative data (e.g., ""reduced infestation"" without quantifiable metrics).

5. Experimental Design: Included: Lab: Controlled conditions (temperature, RH, photoperiod specified). Greenhouse/Screenhouse: Semi-controlled with host plants. Field: Untreated controls + replicated plots (≥3). Clear statistical methods |

| | | (ANOVA, t-tests with variability measures). Excluded: Studies with <3 replicates, unclear methods, or missing sample sizes.

Exclusion Criteria:
Non-Applicable Study Types: Molecular studies (e.g., PCR-based detection) without efficacy data. Formulation chemistry papers without field trials.

Geographic Irrelevance: Studies not conducted in the focal country or with non-native species/strains (unless locally validated).

Incomplete Data: Missing raw mortality rates, sample sizes, or statistical variability. Studies with overlapping datasets (duplicate publications counted once).

Non-Target Focus: Papers solely on non-target effects, pollination, or soil health without pest reduction data.

Scale/Design Flaws: Field trials without controls or with pseudo-replication. Studies reporting only ""presence/absence"" of the focal pest. |